\documentclass[10pt,twocolumn,letterpaper]{article}

\usepackage{cvpr}              
\definecolor{cvprblue}{rgb}{0.21,0.49,0.74}
\usepackage[pagebackref,breaklinks,colorlinks,allcolors=cvprblue]{hyperref}

\usepackage{amsmath,amssymb,amsfonts}
\usepackage{graphicx}
\usepackage{textcomp}
\usepackage[ruled]{algorithm2e}
\usepackage{threeparttable}
\usepackage{multirow}
\usepackage[table]{xcolor}
\usepackage{enumitem}
\usepackage{makecell}
\usepackage[normalem]{ulem}
\usepackage{pifont}

\def\paperID{XXXXX} 
\def\confName{CVPR}
\def\confYear{2026}

\title{Duala: Dual-Level Alignment of Subjects and Stimuli\\ for Cross-Subject fMRI Decoding}

\author{Shumeng Li$^{1,2}$ \; Jintao Guo$^{1,2}$ \; Jian Zhang$^{1,2}$ \; Yulin Zhou$^{1,2}$ \; Luyang Cao$^{1,2}$ \; Yinghuan Shi$^{1,2}$\thanks{Corresponding author: Yinghuan Shi (syh@nju.edu.cn). This work was supported by NSFC Project (62506162, 62506005), Jiangsu Science and Technology Project (BK20251241), and the Fundamental Research Funds for the Central Universities (KG202508).} \\
$^{1}$ State Key Laboratory of Novel Software Technology, Nanjing University\\
$^{2}$ Institute of Brain-Machine Interface, Nanjing University\\
}

\begin{document}
\maketitle
\begin{abstract}
Cross-subject visual decoding aims to reconstruct visual experiences from brain activity across individuals, enabling more scalable and practical brain-computer interfaces. However, existing methods often suffer from degraded performance when adapting to new subjects with limited data, as they struggle to preserve both the semantic consistency of stimuli and the alignment of brain responses.
To address these challenges, we propose \textbf{Duala}, a \textbf{dual-level alignment} framework designed to achieve stimulus-level consistency and subject-level alignment in fMRI-based cross-subject visual decoding.
(1) At the \textbf{stimulus level}, Duala introduces a semantic alignment and relational consistency strategy that preserves intra-class similarity and inter-class separability, maintaining clear semantic boundaries during adaptation. (2) At the \textbf{subject level}, a distribution-based feature perturbation mechanism is developed to capture both global and subject-specific variations, enabling adaptation to individual neural representations without overfitting.
Experiments on the Natural Scenes Dataset (NSD) demonstrate that Duala effectively improves alignment across subjects. Remarkably, even when fine-tuned with only about one hour of fMRI data, Duala achieves over 81.1\% image-to-brain retrieval accuracy and consistently outperforms existing fine-tuning strategies in both retrieval and reconstruction.
Our code is available at {\color{magenta}https://github.com/ShumengLI/Duala}. 
\end{abstract}
    
\section{Introduction}
\label{sec:intro}

\begin{figure}[t]
   \centering
   \includegraphics[width=1.0\linewidth]{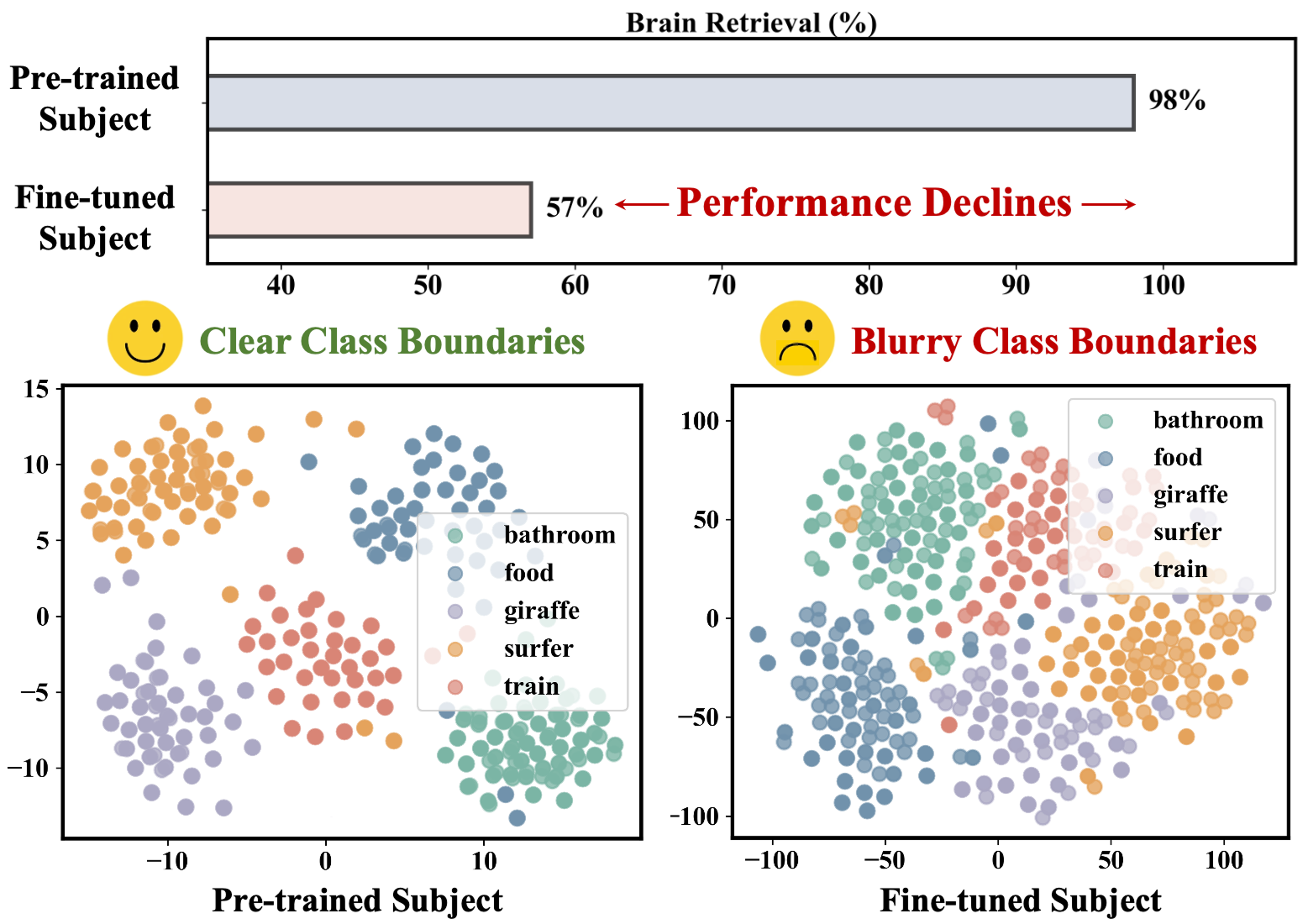}
   \caption{Effectiveness of fine-tuning on retrieval performance and t-SNE visualization. The pre-trained model achieves a high retrieval performance, indicating strong generalization to unseen stimuli. However, after fine-tuning on data from a new subject, the retrieval performance drops significantly. Additionally, the t-SNE visualization of the pre-trained model shows clear class boundaries for the subject’s stimuli, while the t-SNE of the fine-tuned model on the new subject reveals blurry class boundaries, indicating that \textit{the fine-tuning process does not preserve the semantic structure as effectively for the new subject}.}
   \label{fig: intro1}
\end{figure}

Human visual perception is a remarkable process that reflects both the external world and the brain’s internal representations~\cite{dicarlo2012does,bullmore2009complex}. Understanding how the brain encodes and reconstructs visual experiences offers insights into perceptual mechanisms and paves the way for brain-inspired computational models and brain-computer interface applications~\cite{yamins2016using}.
Among the available neural measurement techniques, functional magnetic resonance imaging (fMRI) stands out for its noninvasive nature and high spatial resolution~\cite{fan2022caca}, allowing detailed observation of the cortical activity patterns that underlie visual processing. Consequently, fMRI-based visual decoding~\cite{cox2003functional,kay2008identifying,naselaris2011encoding} has emerged as a vital research direction for uncovering how the human brain interprets and reconstructs visual stimuli.

With the progress of cross-modal foundation models like CLIP~\cite{radford2021learning} and Stable Diffusion~\cite{rombach2022high}, recent studies have achieved remarkable results in reconstructing visual stimuli from brain activity~\cite{seeliger2018generative,shen2019deep,chen2023seeing,lin2022mind,takagi2023high,mai2023unibrain,zeng2024controllable,ozcelik2023natural}. Nevertheless, most of these advances have focused on the single-subject paradigm, where a dedicated decoder is trained for each subject. Due to substantial inter-individual differences in cortical anatomy and cognitive patterns~\cite{wang2020idiosyncratic,quan2024psychometry}, models trained on one subject often fail to generalize to others, limiting their applicability in cross-subject brain decoding~\cite{mai2023unibrain}.
Meanwhile, acquiring extensive fMRI data for each new subject remains costly and time-consuming. For instance, in the Natural Scenes Dataset (NSD)~\cite{allen2022massive}, acquiring high-quality fMRI data for a single subject requires 40 hours of scanning, making large-scale data collection extremely demanding.
Therefore, \textit{how to leverage pre-trained decoding models for effective adaptation to limited data from a new subject} becomes a critical problem.

\begin{figure}[t]
   \centering
   \includegraphics[width=1.0\linewidth]{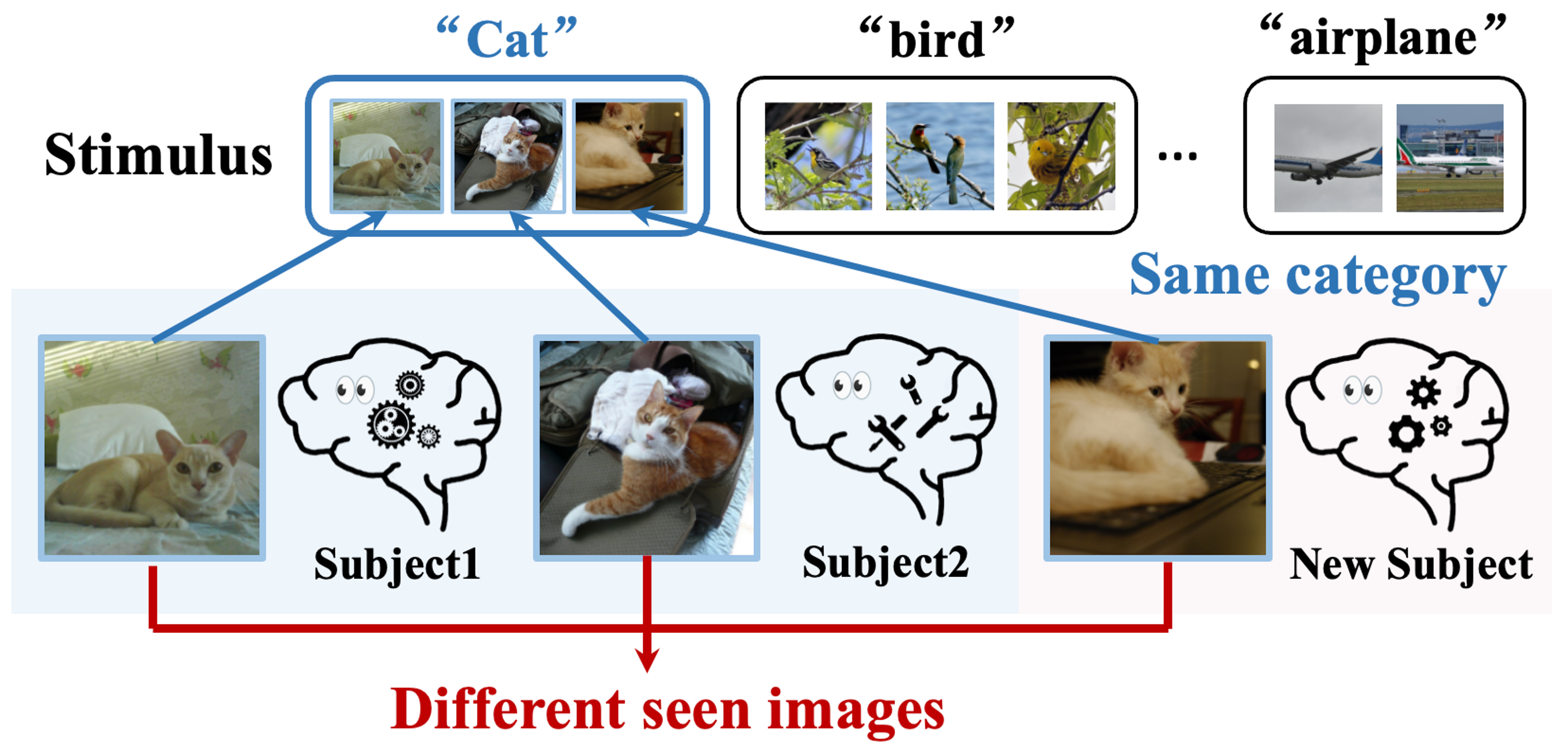}
   \caption{Different subjects are presented with different visual stimuli, even when the stimuli belong to the same category. For example, the subjects are shown an image of a cat, but the actual photos they see differ. }
   \label{fig: intro2}
   \vspace{-5pt}
\end{figure}

Cross-subject visual decoding has emerged as a promising direction to alleviate the limitations of single-subject models. 
Existing approaches~\cite{scotti2024mindeye2,gong2025mindtuner,mai2023unibrain,jung2025mindlink} typically align a new subject’s fMRI representations with those of previously trained subjects by projecting them into a shared representational space. 
However, we observed that while these methods work well during pre-training, they face challenges when applied to new subjects. For instance, in Figure~\ref{fig: intro1}, although the pre-trained model achieves strong retrieval performance on MindEye2~\cite{scotti2024mindeye2}, fine-tuning on new subjects results in a 41\% decrease in image-to-brain retrieval performance.
This decline indicates, a critical yet easily ignored phenomenon that the model struggles to identify the corresponding brain responses from their image embeddings, suggesting that \textit{the learned robust correspondence between brain activity and visual representations at the pre-training stage has been destroyed during the fine-tuning stage}.
This degradation in performance suggests two key challenges.
(1) \textbf{Stimulus-level inconsistency}. As shown in Figure~\ref{fig: intro1}, we observe that for the pre-trained subject, the class boundaries between stimuli are clearly distinguishable, while the t-SNE visualization of the fine-tuned subject reveals blurred category boundaries. 
This indicates that the finetuned model cannot effectively represent  the different stimuli of the new subject.
(2) \textbf{Subject-level misalignment}. As illustrated in Figure~\ref{fig: intro2}, we further observe that an individual image is primarily aligned with only one subject’s fMRI signals, rather than being shared across subjects. In Natural Scenes Dataset (NSD)~\cite{allen2022massive}, more than 90\% of the visual stimuli differ across subjects~\cite{ferrante2024through}, meaning that direct one-to-one alignment is rarely feasible. Consequently, the model struggles to establish consistent inter-subject correspondence, limiting its ability to generalize decoding knowledge to unseen individuals.

To address these challenges, we propose an innovative Dual-level Alignment cross-subject decoding framework, \textbf{Duala}, that operates on both stimulus-level consistency and subject-level alignment. 
At the stimulus level, our design of \textit{Stimulus-level Semantic Preservation} maintains the semantic structure of visual representations. We use a semantic alignment loss to encourage intra-subject discriminability and a relational consistency loss to preserve inter-subject class relationships. 
At the subject level, we introduce \textit{Subject-level Distribution Perturbation}, a feature perturbation strategy designed to capture both global and subject-specific variations. During fine-tuning, the covariance-driven perturbations are applied to the new subject’s representations, helping the model adapt to individual differences while maintaining alignment with the pre-trained feature distributions.
By combining stimulus-level consistency with subject-level alignment, Duala preserves semantic structure while adapting to individual differences, leading to robust and generalizable cross-subject decoding.

In summary, our contributions are as follows:
\begin{itemize}
    \item We propose a novel fine-tuning approach that simultaneously addresses stimulus-level and subject-level alignment to improve cross-subject decoding with limited data.
    \item We introduce a semantic preservation strategy at the stimulus-level to preserve relational consistency among different semantic class.
    \item We develop a distribution perturbation mechanism at the subject-level that adapts the model to the unique brain responses of each new subject.
\end{itemize}
We conduct comprehensive experiments to demonstrate the potential of our Duala, which outperforms state-of-the-art methods for visual decoding on NSD dataset, achieving over 81.1\% retrieval accuracy in image-to-brain matching.

\section{Related Work}
\label{sec:related}

\begin{figure*}[t]
   \centering
   \includegraphics[width=0.93\linewidth]{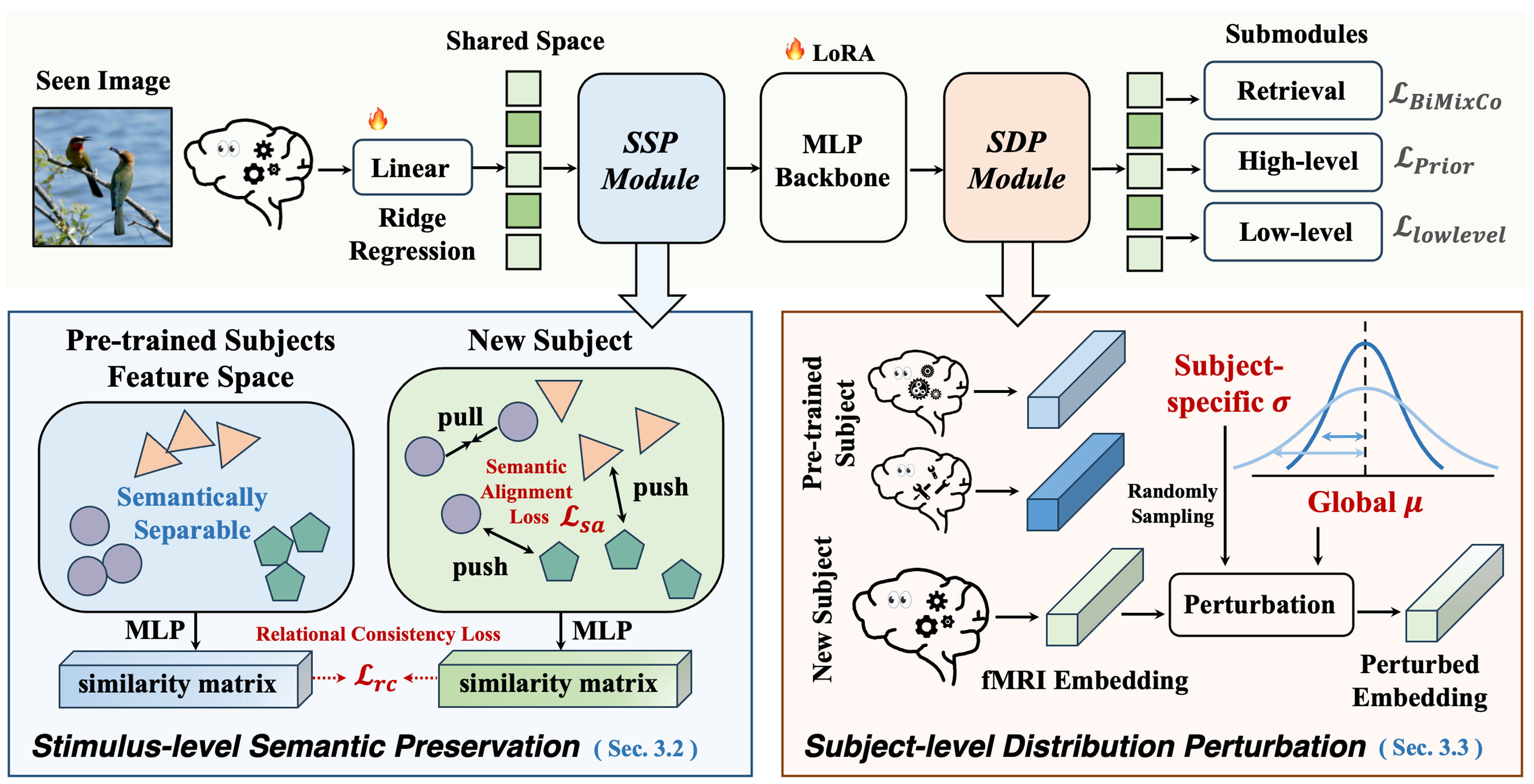}
   \caption{Overview of our Duala framework. Stimulus-level Semantic Preservation maintains the semantic structure of visual representations, and Subject-level Distribution Perturbation enhances cross-subject adaptability.}
   \label{fig: method}
\end{figure*}

\subsection{fMRI-Based Visual Decoding}

fMRI-based visual decoding aims to reconstruct or interpret perceived visual stimuli from neural activity, offering insights into how visual information is represented in the brain~\cite{kay2008identifying,naselaris2011encoding}. Early studies established voxel-wise encoding–decoding frameworks linking neural responses with visual features, which later evolved into deep neural network (DNN)-based approaches that leverage hierarchical feature representations corresponding to cortical organization~\cite{horikawa2017generic}. 
Recent advances have shifted the decoding paradigm from discriminative to generative modeling, where neural signals are projected into latent spaces of large pre-trained models such as VAEs, GANs, or latent diffusion models (LDMs)~\cite{seeliger2018generative,shen2019deep,chen2023seeing,lin2022mind,takagi2023high,mai2023unibrain,zeng2024controllable,ozcelik2023natural,tian2025brainguard}. These latent-space mappings exploit strong priors on semantics and structure, leading to reconstructions that are perceptually coherent and consistent with natural image manifolds. 
However, due to substantial inter-subject variability in brain anatomy and neural responses, most existing methods are trained as subject-specific decoders, each tailored to one subject, which limits their generalizability across subjects.

\subsection{Cross-Subject Functional Alignment}

Individual differences in brain structure and functional organization hinder the generalization of decoding models across subjects. To overcome this, previous alignment methods rely on paired responses to identical visual stimuli, learning mappings through regression or optimal transport objectives~\cite{ferrante2024through,bazeille2019local,thual2023aligning}. These approaches achieve fine-grained alignment but require strictly shared stimuli, which limits their scalability to large datasets such as NSD~\cite{allen2022massive}, where subjects view distinct image sets. 
Recently, representation-based alignment methods project individual brain responses into a shared latent space \cite{dai2025mindaligner,scotti2023reconstructing,scotti2024mindeye2,wang2024mindbridge,bao2025wills,han2024mindformer}. 
MindBridge~\cite{wang2024mindbridge} synthesizes pseudo-shared stimuli to bridge inter-subject gaps. MindEye2~\cite{scotti2024mindeye2} employs ridge regression to project subject-specific signals into a unified latent space for shared decoding, and MindTuner~\cite{gong2025mindtuner} models nonlinear inter-subject relationships. 
While latent-space alignment facilitates cross-subject decoding without shared stimuli, it often neglects stimulus-level distinctions and can distort class structures for new subjects.
In contrast, our approach incorporates both stimulus-level consistency and subject-level alignment, helping to preserve semantic organization while still accommodating individual differences.

\section{Method}
\label{sec:method}
In this work, we propose Duala, a cross-subject decoding framework that integrates stimulus-level consistency and subject-level alignment into a unified learning paradigm. The overall framework is illustrated in Figure~\ref{fig: method}. Specifically, we leverage Stimulus-level Semantic Preservation (SSP Module, in Section~\ref{sec: SSP}) to retain intra- and inter-subject semantic consistency. We further present Subject-level Distribution Perturbation (SDP Module, in Section~\ref{sec: SDP}) to enhance adaptability by aligning new-subject representations with pre-trained feature distributions.

\subsection{Preliminary}

\textbf{Problem Definition.}
In cross-subject visual decoding, the goal is to transfer a pre-trained decoding model from existing subjects to a new subject with limited fMRI data, enabling the reconstruction of visual stimuli from the new subject’s brain activity. Unlike conventional single-subject settings, this problem emphasizes the generalization of learned neural–visual mappings across individuals who exhibit distinct anatomical and functional characteristics.

Formally, let $\mathcal{D}^s = \{(V_i^s, I_i)\}_{i=1}^{M_s}$ denote the fMRI–image pairs collected from subject $s$, where $V_i^s \in \mathbb{R}^{d_s}$ represents voxel responses within selected ROIs and $I_i \in \mathbb{R}^{3 \times H \times W}$ is the corresponding visual stimulus. A decoding model $F_\theta$ is first pre-trained on multiple source subjects $\{s_1, s_2, \dots, s_K\}$ to learn a generalizable mapping between brain activity and visual representations.
When adapting to a new subject $s_N$, only a small amount of fMRI data, $\mathcal{D}^{s_N}$, is available, typically from a single scanning session (about 1 hour of data, 2.5\% of the full dataset). The objective is to adapt the pre-trained model $F_\theta$ to obtain $F_{\theta’}$ that performs reliable decoding for $s_N$: $F_{\theta’}(V^{s_N}) \approx I$.
This formulation defines the data-limited cross-subject adaptation problem that our method aims to address.

\begin{figure}[t]
   \centering
   \includegraphics[width=1.0\linewidth]{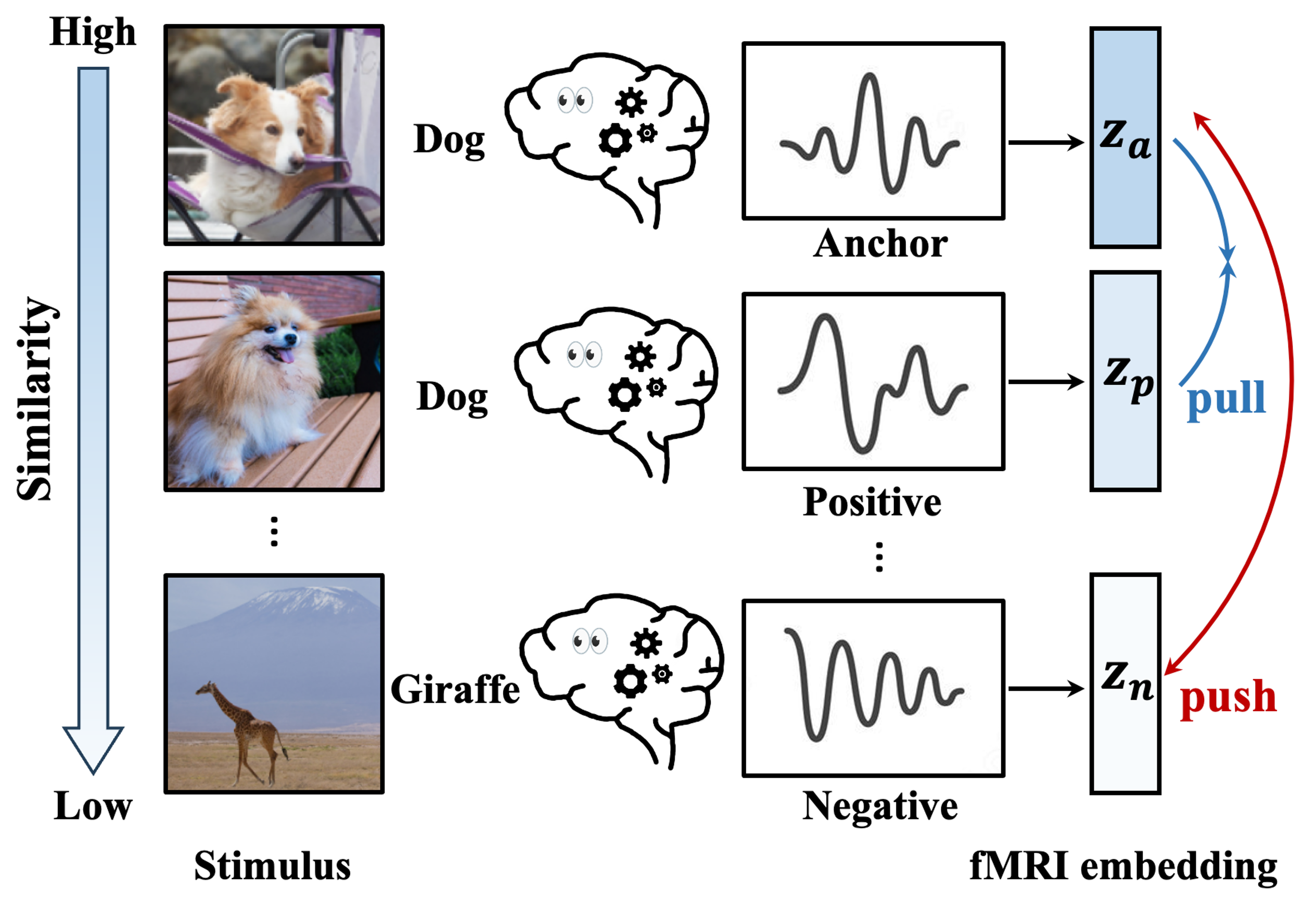}
   \caption{Semantic Alignment Constraint. We sample an anchor and a positive fMRI response elicited by images from the same category (Dog), and a negative from a different category (Giraffe). It encourages fMRI representations of the same stimulus class to be more similar than those of different classes within a subject, preserving intra-class discriminability.}
   \label{fig: stimulus1}
\end{figure}

\textbf{Cross-subject Brain Decoding Baseline.}
As a baseline for cross-subject visual decoding, we adopt a pre-trained model from MindEye2~\cite{scotti2024mindeye2}. To alleviate inter-subject variability, voxel responses from multiple source subjects $\{s_1, \dots, s_K\}$ are first mapped into a shared latent space using ridge regression.
Specifically, subject-specific fMRI voxels $V^{s}$ ($13,000$ to $18,000$ voxels depending on the subject) are linearly projected into a common embedding space to a 4096-dimensional latent vector.

These shared-subject latents are then processed by an MLP backbone consisting of four residual blocks, which transforms them into the OpenCLIP ViT-bigG/14 image token space ($256 \times 1664$ dimensions).
The resulting backbone embeddings are simultaneously passed into a diffusion prior~\cite{ramesh2022hierarchical} that aligns them with CLIP’s image latent distribution~\cite{fa2026decomposing,li2025text}, as well as two lightweight MLP submodules for low-level reconstruction and image retrieval.

The training of the decoding model optimizes a combination of the diffusion prior loss $\mathcal{L}_{\text{prior}}$, the low-level reconstruction loss $\mathcal{L}_{\text{lowlevel}}$, and a bidirectional contrastive retrieval loss $\mathcal{L}_{\text{BiMixCo}}$. Formally, the overall objective is
\begin{equation}
    \mathcal{L}_{\text{dec}} = \mathcal{L}_{\text{prior}} + \alpha_1 \mathcal{L}_{\text{BiMixCo}} + \alpha_2 \mathcal{L}_{\text{lowlevel}},
\end{equation}
where $\alpha_1$ and $\alpha_2$ balance the contributions of the auxiliary losses.

\subsection{Stimulus-level Semantic Preservation}
\label{sec: SSP}
In SSP module, we leverage Semantic Alignment and Relational Consistency constraints to preserve semantic information at the stimulus level. 

\textbf{Semantic Alignment Loss.}
Given the new subject $s_N$, the goal of stimulus-driven alignment is to preserve the semantic structure of stimuli within the subject’s embedding space. 
For each fMRI sample $V_i^{s_N}$, we first obtain the embeddings $z_i^{s_N}$ in the shared latent space after the ridge regression. 
As illustrated in Figure~\ref{fig: stimulus1}, Stimulus-driven alignment encourages embeddings of the same-category stimuli within the new subject to be closer than embeddings of different categories. 
Specifically, for any anchor embedding $z_a^{s_N}$, we identify a positive sample $z_p^{s_N}$ from the same category ($y_a = y_p$) and a negative sample $z_n^{s_N}$ from a different category ($y_n \neq y_a$). The alignment objective enforces that embeddings of the same category are closer than those of different categories:
\begin{equation}
    d(z_a^{s_N}, z_p^{s_N}) < d(z_a^{s_N}, z_n^{s_N}).
\end{equation}

We replace the distance metric with cosine similarity, leading to the following modified constraint:
\begin{equation}
    s(z_a^{s_N}, z_p^{s_N}) > s(z_a^{s_N}, z_n^{s_N}),
\end{equation}
where $s(\cdot, \cdot)$ denotes cosine similarity measure. 
To enforce this similarity constraint, we define the stimulus-driven alignment loss using a triplet loss formulation.
\begin{equation}
    \mathcal{L}_{\text{sa}} = \sum_{a} \max\Big(0,\; m - s(z_a^{s_N}, z_p^{s_N}) + s(z_a^{s_N}, z_n^{s_N})\Big),
\label{eq:sa_loss}
\end{equation}
where $m>0$ is a margin hyperparameter that maintains a minimum separation between categories. 
By optimizing $\mathcal{L}_{\text{sa}}$, the model learns to organize the subject-specific embeddings into a semantically aligned space, encouraging consistent structure across categories while remaining within-subject.

\begin{figure}[t]
   \centering
   \includegraphics[width=1.0\linewidth]{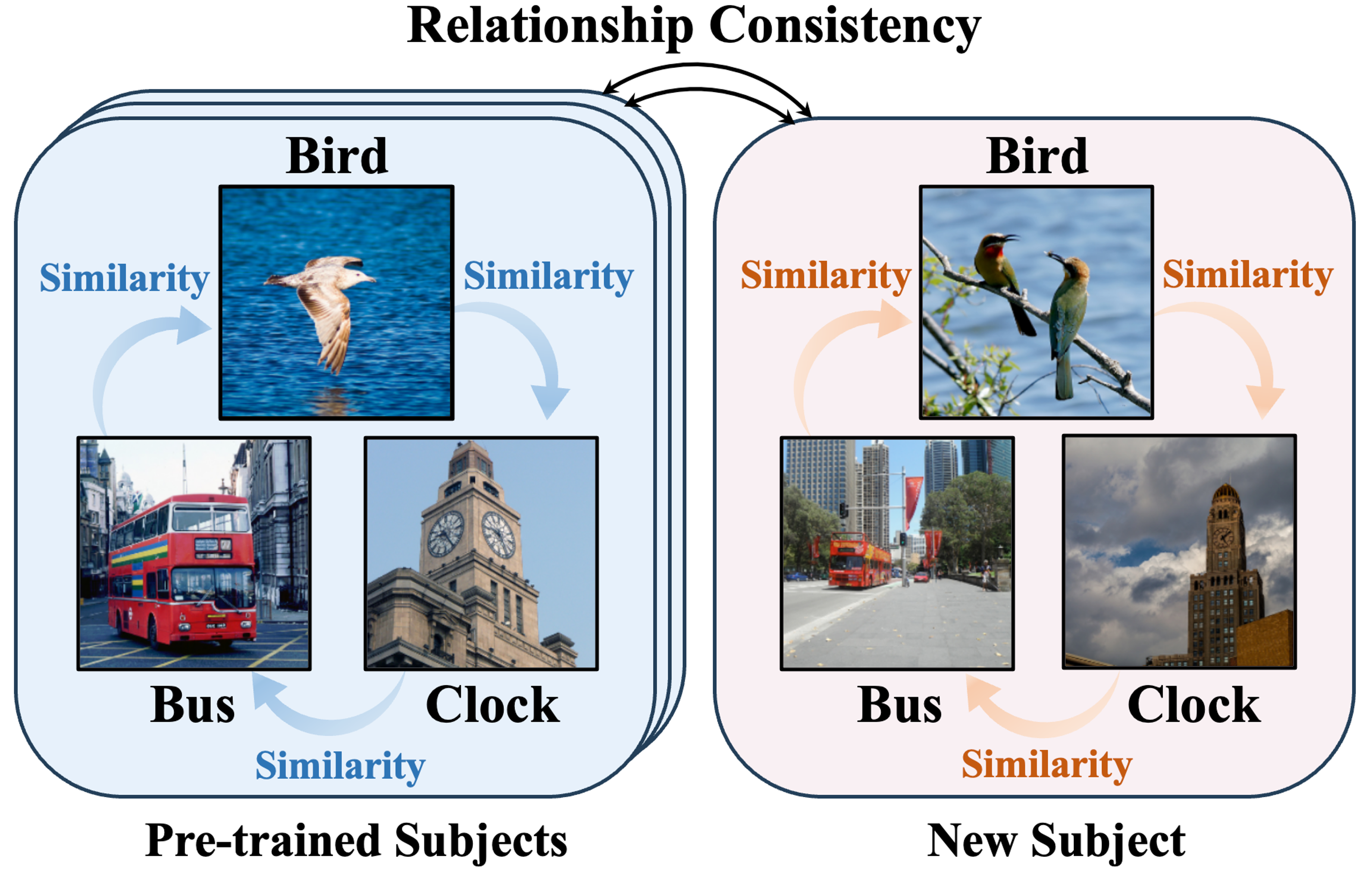}
   \caption{Relational Consistency across subjects. Although subjects view different images even within the same category (\eg, different birds / buses / clocks), the similarity structure among categories should remain consistent across subjects.}
   \label{fig: stimulus2}
\end{figure}

\textbf{Relational Consistency Loss.}
To ensure that the learned brain representations preserve a semantically meaningful structure, our goal is to align the semantic geometry of a new subject with the semantic similarity structure learned during pre-training. To achieve this, we introduce a relational consistency loss objective on the MLP-adapted representations, as shown in Figure~\ref{fig: stimulus2}.

For each subject $s$, we compute the class-level prototype $\mathbf{p}_c^s$ by averaging all normalized embeddings belonging to class $c$. The pairwise cosine similarity between prototypes forms a class-similarity matrix $S^s \in \mathbb{R}^{C \times C}$, where each element is defined as:
\begin{equation}
S^s_{c_1,c_2} = s(\mathbf{p}_{c_1}^s, \mathbf{p}_{c_2}^s),
\end{equation}
where $s(\cdot, \cdot)$ denotes cosine similarity. 
From the pre-trained source subjects $\{s_1, s_2, \dots, s_K\}$, we aggregate these matrices into a reference semantic similarity matrix $S^{\text{ref}}$. During adaptation to a new subject $s_N$, we construct its class-similarity matrix $S^{s_N}$ in the same way and minimize the discrepancy between $S^{s_N}$ and $S^{\text{ref}}$ over valid class pairs:
\begin{equation}
\mathcal{L}_{\mathrm{rc}} = \frac{1}{|\Omega|} 
\sum_{(c_1,c_2)\in\Omega} \big(S^{s_N}_{c_1,c_2} - S^{\text{ref}}_{c_1,c_2}\big)^2,
\end{equation}
where $\Omega$ denotes the set of class pairs for which a reference similarity is available. 
This encourages the adapted model to preserve inter-class similarity patterns inherited from the pre-trained subjects, thereby maintaining semantic consistency in the brain-side feature space.

\subsection{Subject-level Distribution Perturbation}
\label{sec: SDP}
To enable effective cross-subject transfer, we decompose the fMRI representation into a stimulus-driven factor that reflects shared semantic responses, and a subject-specific factor that captures idiosyncratic anatomical and functional variations.

To further improve adaptation to a new subject $s_N$, we introduce a distribution-based feature perturbation strategy grounded in this decomposition. 
We leverage the pre-trained source subjects $\{s_1, \dots, s_K\}$ to model the category-wise distribution of embeddings. 
For each category $c$, we compute the category-level mean $\mu_c$ across all source subjects and the subject-specific deviations $\sigma_c^{s}$:
\begin{equation}
    \mu_c = \frac{1}{K} \sum_{s=1}^K \bar{z}_c^s, \qquad \sigma_c^{s} = \sqrt{\mathrm{Var}(\bar{z}_c^s)},
\end{equation}
where $\bar{z}_c^s$ denotes the average embedding of all samples from category $c$ for subject $s$. Here, $\mu_c$ serves as a rough approximation of the shared stimulus factor $f_{\text{stimulus}}$, while $\sigma_c^{s}$ captures the subject-specific variability.

During adaptation to a new subject $s_N$, we first center each embedding by the category mean: $z_i^{s_N} - \mu_c$, isolating the subject-specific factor. 
We then apply Gaussian perturbations using the source subject deviations $\{\sigma_c^s\}_{s=1}^K$ to augment the subject-specific factor, effectively simulating plausible variations observed across individuals:
\begin{equation}
    \tilde{z}_i^{s_N} = \mu_c + \frac{1}{K} \sum_{s=1}^K \sigma_c^s \odot (z_i^{s_N} - \mu_c),
\end{equation}
where $\odot$ denotes element-wise scaling. 

This distribution-based augmentation preserves the semantic structure provided by the stimulus factor while encouraging the model to be robust to individual-specific variations, facilitating smoother adaptation to the new subject.

Leveraging this decomposition, we further design two complementary constraints to guide model adaptation for a new subject. 
The first constraint operates on the stimulus-driven factor, encouraging embeddings of the same stimulus category within the same subject to cluster closely while pushing apart embeddings from different categories. 
The second constraint focuses on the subject-specific factor, preserving intra-subject consistency and maintaining inter-subject differentiation, thereby preventing the new subject’s unique characteristics from being washed out by naive alignment.
For a new subject $s_N$ and its fMRI embeddings $z_N$, we optimize two objectives, stimulus-driven alignment and subject-specific preservation. 

\subsection{Training Objective}

The overall training objective combines the stimulus-driven alignment loss $\mathcal{L}_{\text{sa}}$ and the subject-specific preservation loss $\mathcal{L}_{\text{rc}}$ to adapt the pre-trained decoding model $F_{\theta}$ to a new subject $s_N$. 
The combined loss is defined as:
\begin{equation}
    \mathcal{L}_{\text{ft}} = \mathcal{L}_{\text{dec}} + \lambda_1 \mathcal{L}_{\text{sa}} + \lambda_2 \mathcal{L}_{\text{rc}},
\end{equation}
where $\lambda_1$ and $\lambda_2$ are hyperparameters that balance the contributions of these objectives.
Optimizing $\mathcal{L}_{\text{ft}}$ encourages the model to organize the new subject's embeddings into a semantically structured space that preserves category-level relationships while maintaining subject-specific characteristics.
This allows the pre-trained decoder to generalize effectively under the data-limited setting, using only a single scanning session from the new subject.

\begin{table*}[t]
\centering
\caption{Quantitative comparison of fMRI-to-image models. Results averaged across subjects 1, 2, 5, and 7 from the Natural Scenes Dataset with 1 hour of data. \textbf{Bold} indicates the best performance.}
\label{tab: main_result}
\scalebox{0.8}{
\begin{tabular}{ccc|cccc|cccc|cc}
\toprule
\multirow{2}{*}{Method} & \multirow{2}{*}{Venue} & \multirow{2}{*}{Subject} & \multicolumn{4}{c}{Low-Level} & \multicolumn{4}{|c}{High-Level} & \multicolumn{2}{|c}{Retrieval} \\
\cmidrule(lr){4-7} \cmidrule(lr){8-11} \cmidrule(lr){12-13}
&&& PixCorr$\uparrow$ & SSIM$\uparrow$ & Alex(2)$\uparrow$ & Alex(5)$\uparrow$ 
& Incep$\uparrow$ & CLIP$\uparrow$ & Eff.$\downarrow$ & SwAV$\downarrow$ 
& Image$\uparrow$ & Brain$\uparrow$ \\
\midrule
MindEye2\scalebox{0.7}{~\cite{scotti2024mindeye2}} & ICML'24 & Avg & 0.195 & 0.419 & 84.2\% & 90.6\% & 81.2\% & 79.2\% & 0.810 & 0.468 & 79.0\% & 57.4\% \\
MindAligner\scalebox{0.7}{~\cite{dai2025mindaligner}} & ICML'25 & Avg & 0.206 & 0.414 & 85.6\% & 91.6\% & 81.1\% & 82.0\% & 0.802 & 0.463 & 79.0\% & 75.3\% \\
MindTuner\scalebox{0.7}{~\cite{gong2025mindtuner}} & AAAI'25 & Avg & 0.224 & \textbf{0.420} & 87.8\% & \textbf{93.6\%} & 84.8\% & 83.5\% & \textbf{0.780} & \textbf{0.440} & 83.1\% & 76.0\% \\
\textbf{Duala \scalebox{0.7}{(Ours)}} & - & Avg & \textbf{0.230} & 0.416 & \textbf{87.9\%} & 93.5\% & \textbf{85.4\%} & \textbf{83.5\%} & 0.781 & 0.445 & \textbf{84.5\%} & \textbf{81.1\%} \\
\midrule
MindEye2\scalebox{0.7}{~\cite{scotti2024mindeye2}} & ICML'24 & Subj1 & 0.235 & \textbf{0.428} & 88.02\% & 93.33\% & 83.56\% & 81.76\% & 0.798 & 0.459 & 93.96\% & 77.63\% \\
MindAligner\scalebox{0.7}{~\cite{dai2025mindaligner}} & ICML'25 & Subj1 & 0.226 & 0.415 & 88.19\% & 93.26\% & 83.48\% & 81.76\% & 0.800 & 0.459 & 90.90\% & 86.88\% \\
MindTuner\scalebox{0.7}{~\cite{gong2025mindtuner}} & AAAI'25 & Subj1 & \textbf{0.262} & 0.422 & 90.60\% & 94.90\% & 85.80\% & 84.60\% & 0.774 & 0.433 & 94.20\% & 87.40\% \\
\textbf{Duala \scalebox{0.7}{(Ours)}} & - & Subj1 & 0.253 & 0.418 & \textbf{90.81\%} & \textbf{95.29\%} & \textbf{86.62\%} & \textbf{85.11\%} & \textbf{0.768} & \textbf{0.433} & \textbf{94.77\%} & \textbf{91.22\%} \\
\midrule
MindEye2\scalebox{0.7}{~\cite{scotti2024mindeye2}} & ICML'24 & Subj2 & 0.200 & \textbf{0.433} & 85.00\% & 92.13\% & 81.86\% & 79.89\% & 0.807 & 0.454 & 90.53\% & 67.18\% \\
MindAligner\scalebox{0.7}{~\cite{dai2025mindaligner}} & ICML'25 & Subj2 & 0.218 & 0.426 & 88.08\% & 93.33\% & 82.47\% & 81.62\% & 0.791 & 0.452 & 90.04\% & 85.61\% \\
MindTuner\scalebox{0.7}{~\cite{gong2025mindtuner}} & AAAI'25 & Subj2 & 0.225 & 0.425 & 89.10\% & \textbf{95.10\%} & 84.80\% & 83.70\% & 0.781 & \textbf{0.440} & 93.70\% & 82.80\% \\
\textbf{Duala \scalebox{0.7}{(Ours)}} & - & Subj2 & \textbf{0.236} & 0.428 & \textbf{89.21\%} & 94.35\% & \textbf{85.35\%} & \textbf{83.74\%} & \textbf{0.778} & 0.443 & \textbf{94.64\%} & \textbf{91.47\%} \\
\midrule
MindEye2\scalebox{0.7}{~\cite{scotti2024mindeye2}} & ICML'24 & Subj5 & 0.175 & 0.405 & 83.11\% & 90.83\% & 82.32\% & 78.53\% & 0.781 & 0.444 & 66.94\% & 46.96\% \\
MindAligner & ICML'25 & Subj5 & 0.197 & 0.409 & 84.69\% & 91.61\% & 82.63\% & 80.12\% & 0.784 & 0.454 & 70.62\% & 65.95\% \\
MindTuner\scalebox{0.7}{~\cite{gong2025mindtuner}} & AAAI'25 & Subj5 & 0.208 & \textbf{0.415} & \textbf{86.80\%} & \textbf{93.70\%} & \textbf{87.70\%} & \textbf{85.90\%} & \textbf{0.750} & \textbf{0.422} & 72.20\% & 68.10\% \\
\textbf{Duala \scalebox{0.7}{(Ours)}} & - & Subj5 & \textbf{0.214} & 0.413 & 86.50\% & 93.07\% & 86.36\% & 84.46\% & 0.772 & 0.439 & \textbf{76.59\%} & \textbf{71.20\%} \\
\midrule
MindEye2\scalebox{0.7}{~\cite{scotti2024mindeye2}} & ICML'24 & Subj7 & 0.170 & 0.408 & 80.70\% & 85.90\% & 74.90\% & 79.84\% & 0.834 & 0.504 & 64.44\% & 37.77\% \\
MindAligner\scalebox{0.7}{~\cite{dai2025mindaligner}} & ICML'25 & Subj7 & 0.183 & 0.407 & 81.45\% & 88.31\% & 79.92\% & \textbf{80.83\%} & 0.834 & 0.487 & 64.18\% & 62.58\% \\
MindTuner\scalebox{0.7}{~\cite{gong2025mindtuner}} & AAAI'25 & Subj7 & 0.202 & \textbf{0.417} & 84.50\% & 90.80\% & 80.70\% & 79.60\% & 0.817 & 0.465 & 71.90\% & 65.50\% \\
\textbf{Duala \scalebox{0.7}{(Ours)}} & - & Subj7 & \textbf{0.215} & 0.405 & \textbf{85.17\%} & \textbf{91.16\%} & \textbf{83.34\%} & 80.72\% & \textbf{0.805} & \textbf{0.463} & \textbf{71.94\%} & \textbf{70.53\%} \\
\bottomrule
\end{tabular}}
\vspace{-3pt}
\end{table*}

\section{Experiments}
\label{sec:exp}

\subsection{Dataset}
We conduct our experiments on the Natural Scenes Dataset (NSD)~\cite{allen2022massive}, one of the largest publicly available fMRI datasets for visual decoding. The dataset contains neural responses from multiple subjects viewing complex natural images sampled from the MSCOCO-2017 collection~\cite{lin2014microsoft}. Each participant completed multiple sessions using a 7-T fMRI scanner, with each session comprising 750 trials and lasting approximately one hour. Following the data-limited setup of prior works~\cite{scotti2024mindeye2}, we use only a single session per subject for fine-tuning, enabling evaluation under constrained data conditions. 

\subsection{Implementation Details}
Our framework is implemented in PyTorch and trained on a single NVIDIA A800 GPU. We adopt the same pre-trained backbone as MindEye2~\cite{scotti2024mindeye2} and keep the diffusion prior and MLP modules frozen. Following the strategy of MindTuner~\cite{gong2025mindtuner}, we introduce LoRA~\cite{hu2022lora} adapters with rank 8 in the MLP backbone for efficient fine-tuning on new subjects. For each new subject, we use a single session of fMRI data (approximately 1 hour) and apply a batch size of 10 as previous work~\cite{gong2025mindtuner}. Optimization is performed using AdamW~\cite{loshchilov2017decoupled} with a learning rate of 3e-4 and OneCycle learning rate schedule~\cite{smith2019super}. Training is conducted for 150 epochs. Additionally, BiMixCo is replaced with the SoftCLIP~\cite{scotti2023reconstructing} loss for one-third of the training iterations. 
For the Stimulus-level Semantic Preservation, we apply the semantic alignment loss and relational consistency Loss, with weighting factors $\lambda_1=1.0$ and $\lambda_2=0.1$, respectively.

\textbf{Metrics.}
To quantitatively evaluate fMRI-to-image reconstruction, we consider both low-level and high-level aspects of the generated images. Low-level metrics capture basic visual properties such as pixel intensity and edge alignment, while high-level metrics reflect semantic content. Following prior works~\cite{scotti2024mindeye2,dai2025mindaligner,gong2025mindtuner}, we adopt PixCorr, SSIM~\cite{wang2004image}, AlexNet(2), and AlexNet(5)~\cite{krizhevsky2012imagenet} to evaluate low-level fidelity, and Inception~\cite{szegedy2016rethinking}, CLIP~\cite{radford2021learning}, EfficientNet-B~\cite{tan2019efficientnet}, and SwAV~\cite{caron2020unsupervised} to assess high-level semantic consistency. 

\begin{figure*}[t]
   \centering
   \includegraphics[width=0.9\linewidth]{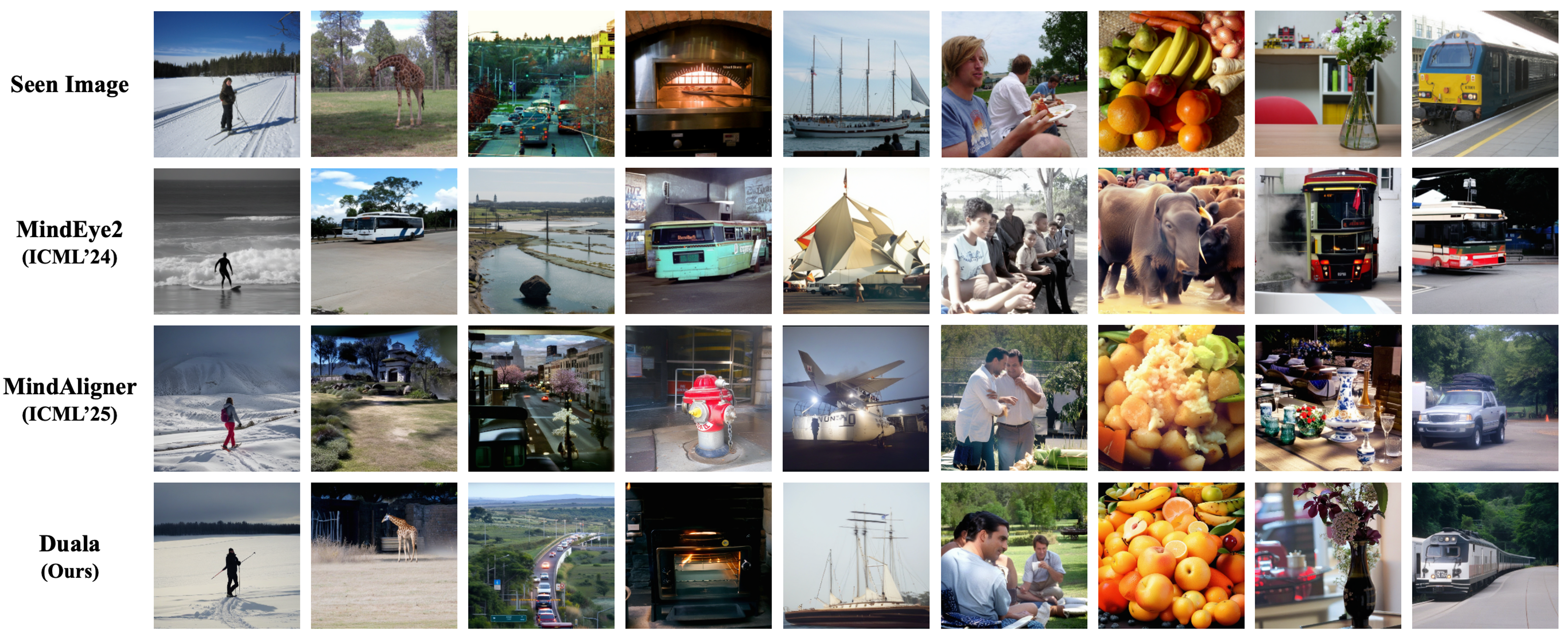}
   \caption{Visualization of reconstruction results with only 1 hour of data. }
   \label{fig: visual}
   \vspace{-5pt}
\end{figure*}

\begin{table}[t]
\centering
\caption{Comparison of training parameters during fine-tuning. Ridge regression parameters correspond to Subject 1. }
\label{tab: params}
\scalebox{0.8}{
\begin{tabular}{c|ccc|c}
\midrule
\multirow{2}{*}{Method} & \multicolumn{3}{c|}{Model Parts} & \multirow{2}{*}{Total} \\
& Ridge & MLP & Prior & \\
\midrule
MindEye2\scalebox{0.7}{~\cite{scotti2024mindeye2}} & 64.41M & 1903M & 260M & 2.2G \\
MindAligner\scalebox{0.7}{~\cite{dai2025mindaligner}} & 139.23M & 0M & 0M & 139.23M \\
MindTuner\scalebox{0.7}{~\cite{gong2025mindtuner}} & 64.41M & 12.3M & 0M & 76.71M \\
\textbf{Duala \scalebox{0.7}{(Ours)}} & \textbf{64.41M} & \textbf{4.68M} & \textbf{0M} & \textbf{69.09M} \\
\bottomrule
\end{tabular}}
\vspace{-5pt}
\end{table}

\subsection{Experimental Results}
\begin{figure}[t]
   \centering
   \includegraphics[width=0.95\linewidth]{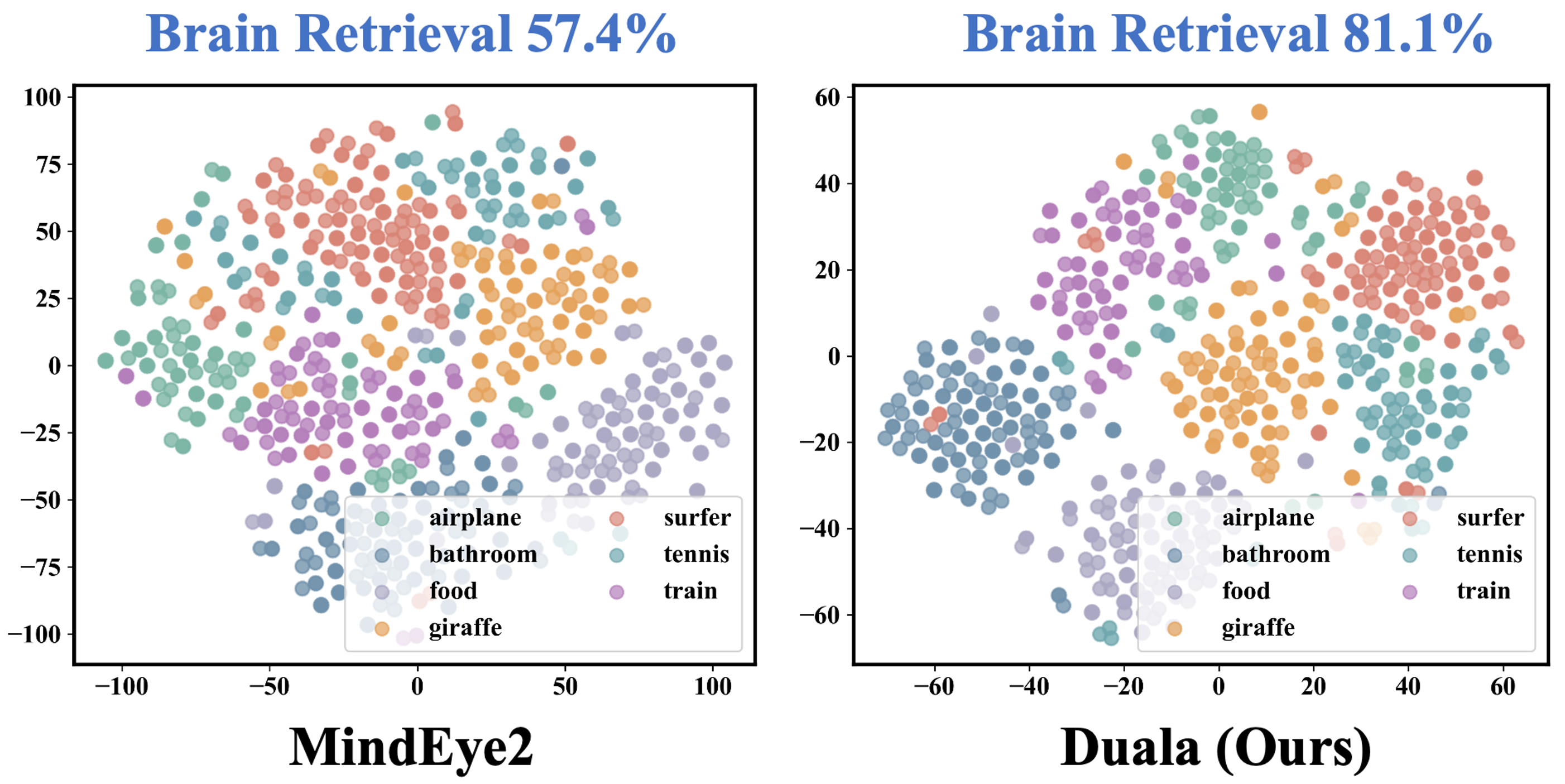}
   \caption{Comparison on t-SNE of MindEye2 and Duala fine-tuned on subject 1. Duala shows clearer class boundaries, indicating improved stimulus-level semantic preservation.}
   \label{fig: tsne}
\end{figure}

\textbf{Image and Brain Retrieval.}
Table~\ref{tab: main_result} reports the quantitative comparison of Duala with recent state-of-the-art cross-subject decoding frameworks, including MindEye2~\cite{scotti2024mindeye2}, MindAligner~\cite{dai2025mindaligner}, and MindTuner~\cite{gong2025mindtuner}.
Across both image retrieval and brain retrieval, Duala achieves consistent improvements under the 1-hour training condition. On average, Duala attains 84.5\% image retrieval accuracy and 81.1\% brain retrieval accuracy, outperforming MindTuner~\cite{gong2025mindtuner} by 1.4\% and 5.1\%, respectively. This improvement highlights the benefit of jointly performing stimulus-level relational consistency and subject-level feature perturbation, which together preserve semantic structure while enhancing generalization to new subjects.
Notably, these gains are consistent across all four NSD subjects (1, 2, 5, and 7), and both forward (brain-to-image) and backward (image-to-brain) retrieval improve for each subject. This indicates that Duala generalizes reliably across individuals and yields stable retrieval.
In addition, we provide a t-SNE comparison of the feature representations for Subject 1 (Figure~\ref{fig: tsne}). While MindEye2~\cite{scotti2024mindeye2} shows considerable overlap between classes after fine-tuning, Duala maintains clearer separations among stimulus classes, confirming the effectiveness of the stimulus-level semantic preservation module in preserving intra-class coherence. 

\textbf{Image Reconstruction.}
In terms of image reconstruction, Duala achieves the highest or second-highest scores across most perceptual and pixel-level metrics. Averaged across subjects, Duala obtains the best results in Pixel Correlation (0.230), AlexNet(2) (87.9\%), Inception (85.4\%), and CLIP (83.5\%), indicating superior preservation of low-level details and high-level semantic fidelity. Compared with MindTuner~\cite{gong2025mindtuner}, Duala improves Inception similarity by 0.6\% and maintains comparable SSIM and AlexNet(5) scores, demonstrating that the proposed dual-level constraints enhance representational quality without sacrificing structural coherence.
Figure~\ref{fig: visual} presents examples of images reconstructed from fMRI signals across multiple subjects. Compared with previous methods~\cite{scotti2024mindeye2,dai2025mindaligner}, our approach produces images that better reflect the correct semantic category of the stimulus, with fewer misclassifications or cross-category confusions. This demonstrates that our method helps maintain clear distinctions between classes leading to more faithful and coherent reconstructions.

\begin{table*}[t]
\centering
\caption{Ablation studies on Duala. Results of subject 1 from NSD Dataset with 1 hour of data. \textbf{Bold} indicates the best performance.}
\label{tab: ablation}
\scalebox{0.8}{
\begin{tabular}{ccc|cccc|cccc|cc}
\toprule
\multicolumn{3}{c|}{Key Components} & \multicolumn{4}{c}{Low-Level} & \multicolumn{4}{|c}{High-Level} & \multicolumn{2}{|c}{Retrieval} \\
\cmidrule(lr){4-7} \cmidrule(lr){8-11} \cmidrule(lr){12-13}
SDP Module & SSP $\mathcal{L}_{\text{sa}}$ & SSP $\mathcal{L}_{\text{rc}}$ & PixCorr$\uparrow$ & SSIM$\uparrow$ & Alex(2)$\uparrow$ & Alex(5)$\uparrow$ 
& Incep$\uparrow$ & CLIP$\uparrow$ & Eff.$\downarrow$ & SwAV$\downarrow$ 
& Image$\uparrow$ & Brain$\uparrow$ \\
\midrule        
\ding{56} & \ding{56} & \ding{56} & 0.243 & 0.418 & 89.33\% & 94.28\% & 84.24\% & 83.35\% & 0.791 & 0.446 & 93.31\% & 89.92\% \\
\ding{52} & \ding{56} & \ding{56} & 0.244 & 0.414 & 89.70\% & 94.34\% & 84.45\% & 83.68\% & 0.779 & 0.441 & 93.84\% & 90.59\% \\
\ding{56} & \ding{52} & \ding{56} & 0.252 & 0.413 & 90.80\% & \textbf{95.47\%} & 86.16\% & 84.00\% & 0.772 & 0.435 & 91.89\% & \textbf{92.38\%} \\
\ding{52} & \ding{52} & \ding{56} & 0.249 & 0.416 & 90.19\% & 94.90\% & 85.33\% & 83.92\% & 0.783 & 0.445 & 93.86\% & 91.43\% \\
\midrule
\ding{52} & \ding{52} & \ding{52} & \textbf{0.253} & \textbf{0.418} & \textbf{90.81\%} & 95.29\% & \textbf{86.62\%} & \textbf{85.11\%} & \textbf{0.768} & \textbf{0.433} & \textbf{94.77\%} & 91.22\% \\
\bottomrule
\end{tabular}}
\end{table*}

\begin{table}[t]
\centering
\caption{Sensitivity of Duala performance to loss weights. Results of subject 1 from the Natural Scenes Dataset with 1 hour of data.}
\label{tab: weights}
\scalebox{0.75}{
\begin{tabular}{cc|cccc|cc}
\toprule
$\lambda_1$ & $\lambda_2$ & PixCorr$\uparrow$ & SSIM$\uparrow$ & Incep$\uparrow$ & CLIP$\uparrow$ & Image$\uparrow$ & Brain$\uparrow$ \\
\midrule
0.5 & 0.05 & 0.250 & 0.417 & 85.87\% & 84.40\% & 93.93\% & 90.94\% \\
1.0 & 0.05 & 0.251 & 0.418 & 85.21\% & 83.74\% & 93.88\% & 91.14\% \\
1.0 & 0.1 & 0.253 & 0.418 & 86.62\% & 85.11\% & 94.77\% & 91.22\% \\
1.0 & 0.5 & 0.253 & 0.417 & 85.77\% & 84.86\% & 94.52\% & 90.87\% \\
\bottomrule
\end{tabular}}
\end{table}

\subsection{Ablation Studies}
\textbf{Effectiveness of Key Components.}
We conducted ablation experiments to evaluate the contribution of each component in Duala, as summarized in Table~\ref{tab: ablation}.
The first row serves as our baseline, which adopts the same LoRA~\cite{hu2022lora} and Skip-LoRA fine-tuning strategy as MindTuner~\cite{gong2025mindtuner} without any of Duala's alignment components. Adding the SDP module alone yields consistent gains in retrieval and reconstruction. Introducing $\mathcal{L}_{\text{sa}}$ of SSP module markedly strengthens semantic separability and pushes brain retrieval to 92.38\%, but slightly reduces image retrieval, suggesting that it tightens class cohesion yet can bias forward matching when used alone. Combining SDP and $\mathcal{L}_{\text{sa}}$ recovers forward matching while maintaining high brain retrieval. Finally, the full model (SDP+$\mathcal{L}_{\text{sa}}$+$\mathcal{L}_{\text{rc}}$) achieves the best overall trade-off, with state-of-the-art high-level semantics, the lowest dissimilarities (Eff. 0.768, SwAV 0.433), and the highest image retrieval, while keeping brain retrieval competitive.
These results validate that Duala effectively integrates both stimulus-level and subject-level regularization, producing superior decoding performance.

\textbf{Sensitivity to Loss Weights.}
Table~\ref{tab: weights} evaluates the influence of the weights of semantic alignment loss ($\lambda_1$) and the relational consistency loss ($\lambda_2$) on Duala’s performance.
It shows that Duala is robust to the semantic alignment weight $\lambda_1$ increasing from $0.5$ to $1.0$, producing only minor changes.
In contrast, performance is more sensitive to the relational consistency weight $\lambda_2$. A light–to–moderate $0.1$ performs best. Setting too high $0.5$ reduces both retrieval metrics, suggesting over-regularization toward source relational geometry that hampers new-subject adaptation. Low-level fidelity (PixCorr and SSIM) remains largely unchanged across settings, indicating that $\lambda_2$ primarily shapes semantic geometry rather than pixel structure. 

\subsection{Further Analysis}
\textbf{Efficiency Analysis.}
Table~\ref{tab: params} compares the number of trainable parameters during fine-tuning. The parameters of the ridge regression layer vary across subjects, and the numbers reported correspond to Subject 1. Duala introduces only 4.68M trainable parameters for its MLP module while keeping the alignment network frozen, resulting in a total of 69.09M parameters, significantly fewer than MindEye2 (2.2G) and MindTuner (76.7M). Despite this lightweight design, Duala achieves the highest decoding performance across all evaluation metrics, demonstrating superior parameter efficiency and fine-tuning stability.

\begin{figure}[t]
   \centering
   \includegraphics[width=0.95\linewidth]{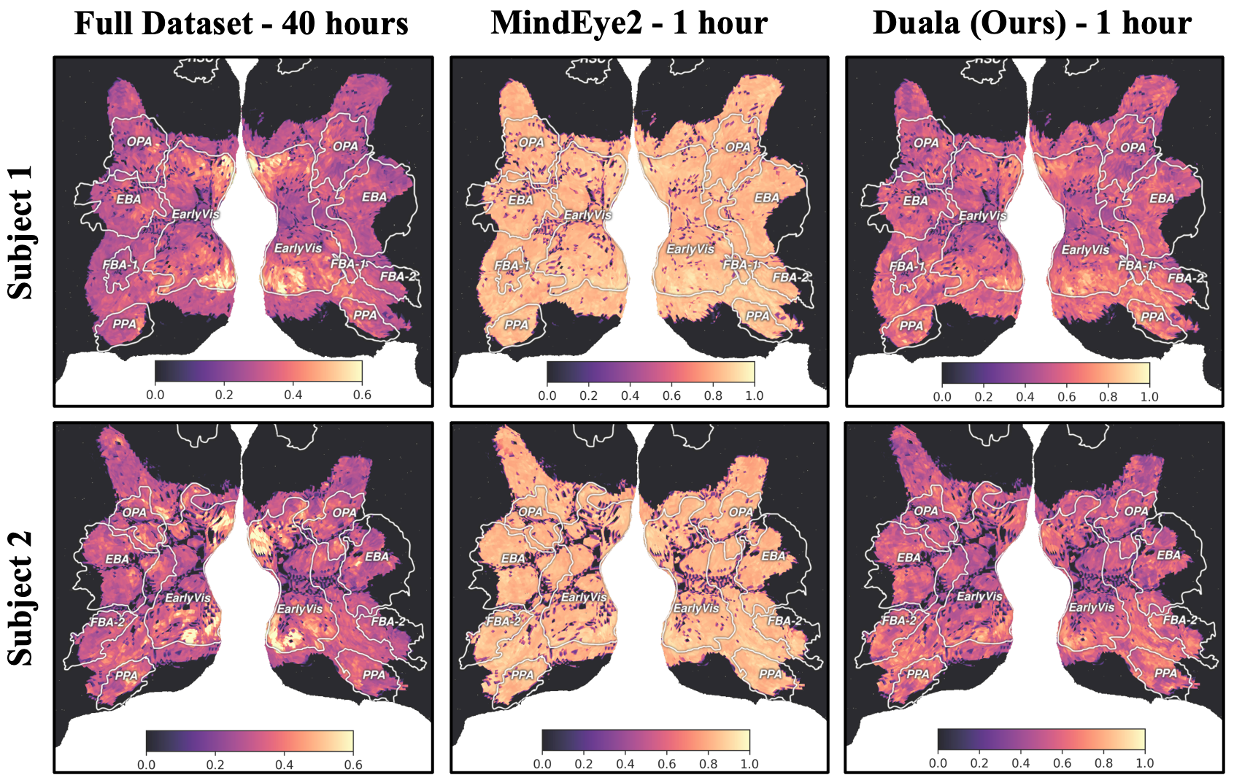}
   \caption{Visualization of transfer quantity in brain heatmaps. It shows that our method produces TQ maps with clear region-specific hotspots that match the 40-hour (full dataset) model.}
   \label{fig: heatmap}
   \vspace{-5pt}
\end{figure}

\textbf{Functional Alignment Analysis.}
To further investigate cross-subject functional alignment, we visualize brain regions associated with stimulus representations using the Transfer Quantity (TQ) metric. Specifically, we compute TQ values from ridge regression. The resulting heatmaps highlight regions with stronger or weaker cross-subject correspondence, providing an intuitive view of inter-individual variability.
In the full-dataset model, high TQ values are concentrated in canonical visual regions (\eg, EarlyVis, OPA, EBA, PPA), reflecting fine-grained functional specialization. By contrast, MindEye2~\cite{scotti2024mindeye2} spreads high TQ values more evenly across the cortex, suggesting that its subject-specific mapping fails to preserve this regional structure.
In addition, our method produces TQ maps with clear region-specific hotspots that match the 40-hour model.


\section{Conclusion}
\label{sec:concl}
In this work, we presented \textbf{Duala}, a fine-tuning framework for cross-subject fMRI-to-image decoding that considers both stimulus-level and subject-level alignment.
By introducing Stimulus-level Semantic Preservation to maintain relational structure across visual classes and Subject-level Distribution Perturbation to adapt to individual brain response variability, Duala effectively leverages limited data from new subjects. Extensive experiments on NSD dataset demonstrate that our approach consistently improves both reconstruction and retrieval performance compared to other methods, highlighting its ability to preserve semantic integrity while accommodating subject-specific differences.
\clearpage
{
    \small
    \bibliographystyle{ieeenat_fullname}
    \bibliography{main}
}


\end{document}